\documentclass{ecai} 

\usepackage{latexsym}
\usepackage{amssymb}
\usepackage{amsmath}
\usepackage{amsthm}
\usepackage{booktabs}
\usepackage{enumitem}
\usepackage{graphicx}
\usepackage{color}
\usepackage{algorithm}
\usepackage{algorithmicx}
\usepackage{algpseudocode}
\usepackage{xcolor}
\usepackage{float}
\usepackage{subfigure}
\usepackage[colorlinks=true, allcolors=blue]{hyperref}

\newcommand{\BibTeX}{B\kern-.05em{\sc i\kern-.025em b}\kern-.08em\TeX}

\begin{document}


\begin{frontmatter}


\paperid{3443} 


\title{Conservative Query and Adaptive Regularization for Offline RL under Uncertainty Estimation}


\author[A]{\fnms{Li-Rong}~\snm{Zhou}\footnote{Equal Contribution.}}
\author[A]{\fnms{Qin-Wen}~\snm{Luo}\footnotemark}
\author[A]{\fnms{Sheng-Jun}~\snm{Huang}\thanks{Corresponding Author. Email: huangsj@nuaa.edu.cn}} 
\address[A]{ Nanjing University of Aeronautics and Astronautics, Nanjing, China \\ \{lrzhou, luoqw8, huangsj\}@nuaa.edu.cn}


\begin{abstract}
Offline reinforcement learning (RL) aims to learn an effective policy from a static dataset, but the achievable performance is fundamentally limited by the coverage of the dataset. The action preference query mechanism leverages expert feedback without requiring environment interaction, enabling performance improvements during offline training while avoiding the cost and risks associated with online fine-tuning. However, existing methods still face significant challenges, both in designing helpful query strategies and in efficiently exploiting the collected preferences. Current approaches typically select queries based solely on the distance between policy actions and dataset actions, and apply naive constraints that compel the policy to remain close to the queried preferences. Such strategies often lead to unstable and inefficient policy updates, and pose challenges for integration with value regularization methods. To address these issues, we propose conservative query and adaptive regularization under uncertainty estimation, a novel and lightweight framework that jointly tackles both the challenges of the preference query and exploitation. Specifically, we first employ the Morse neural network to quantify the uncertainty of the given action relative to the dataset. To facilitate helpful queries, we introduce the uncertainty-driven conservative query mechanism that leverages uncertainty estimation to selectively query actions near the dataset to preserve the stability of Bellman updates. For more effective preference exploitation, we propose the uncertainty-aware adaptive regularization to dynamically modulates the strength of data-level constraints based on the uncertainty of policy actions, enabling the policy to benefit from reliable Bellman updates. We integrate our framework with CQL and perform extensive experiments on the D4RL benchmark. The results demonstrate that our method achieves superior or competitive performance across various tasks.
\end{abstract}

\end{frontmatter}

\section{Introduction}\label{introduction}

In reinforcement learning (RL), agents aim to optimize sequential decision-making by selecting actions that maximize the expected cumulative reward within a given environment \cite{sutton1998introduction}. However, the need for active interaction with the environment renders online RL impractical for many real-world applications, particularly in high-stakes domains such as robotics \cite{singh2022reinforcement}, autonomous driving \cite{kiran2021deep}, and healthcare \cite{tang2022leveraging}, where online data collection is often costly, time-consuming, or unsafe. Offline reinforcement learning (offline RL) addresses this challenge by learning from fixed, pre-collected datasets without additional environment interaction, offering a more practical and scalable alternative \cite{jaques2019way, levine2020offline, fujimoto2018off, ernst2005tree, brandfonbrener2021offline, zhang2020gendice}. Unfortunately, offline RL faces a fundamental challenge due to the distribution shift between the behavior policy that generated the dataset and the target policy being learned. This mismatch frequently results in the selection of out-of-distribution actions, leading to inaccurate value estimates and compounding error propagation during training \cite{kumar2019stabilizing, fujimoto2018off}. To address this issue, the prevailing approach in offline RL constrains the learned policy to remain within the support of the dataset, mitigating the distribution shift through a pessimistic learning principle.

\begin{figure}[t]
\begin{center}
\centerline{\includegraphics[width=\columnwidth]{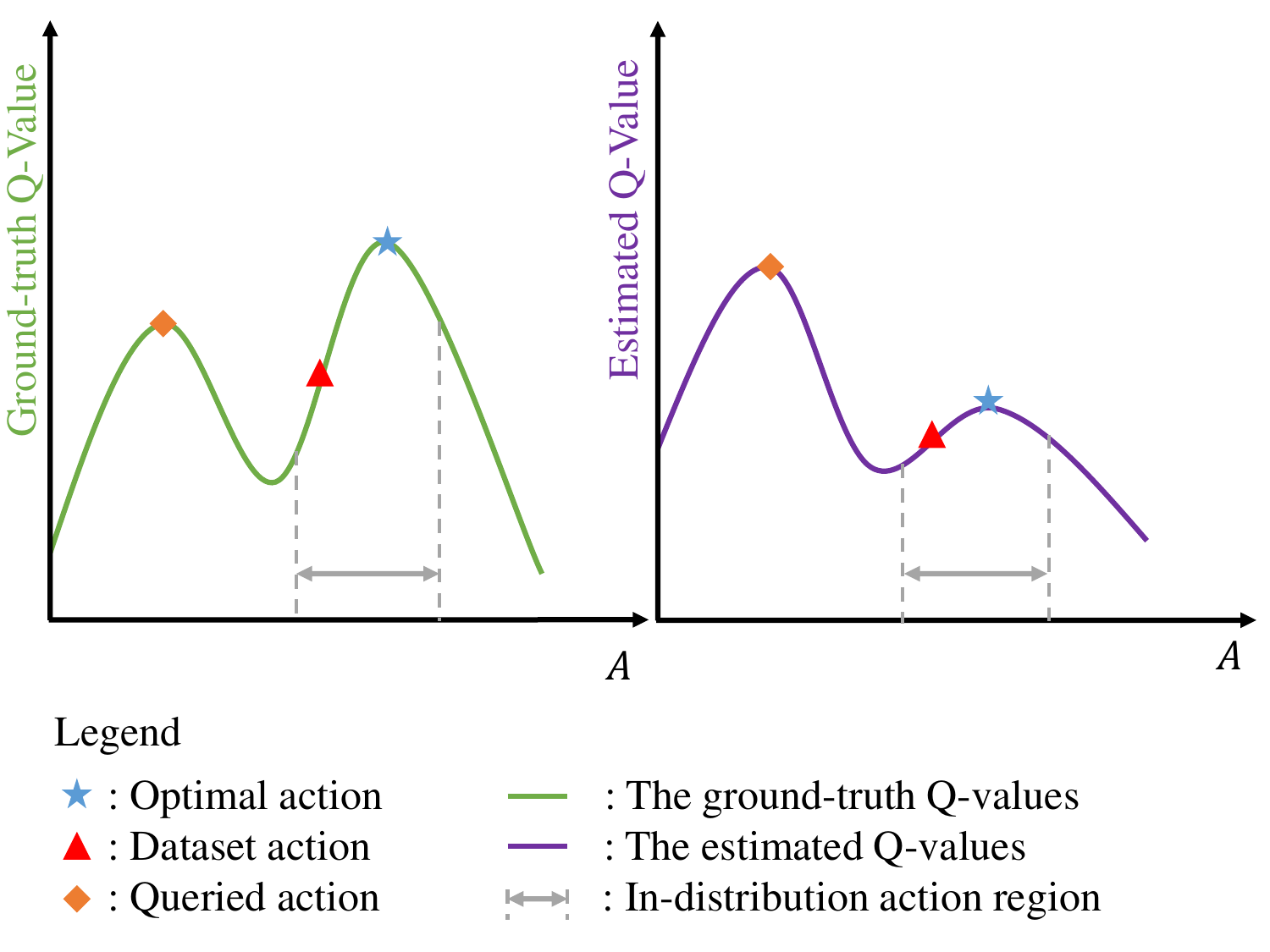}}
\caption{Inaccurate value estimation induced by query shift. (left) The ground-truth Q-values provided by an oracle. (right) The estimated Q-values from the trained critic in value regularization methods.}
\label{fig:motivation}
\end{center}
\end{figure}

Despite significant progress in offline RL, the achievable performance of the learned policy remains fundamentally limited by the data distribution of the fixed dataset. To overcome this limitation, online fine-tuning that initializes from offline results \cite{luo2024optimistic, beeson2022improving, wang2023train} has emerged as a promising direction, leveraging limited online interactions to further improve policy performance. However, in high-risk or cost-sensitive domains, even minimal online interaction could be infeasible. As an alternative, the mechanism for action preference query offers a compelling approach to enhance policy performance without requiring direct interaction with the environment \cite{yang2023boosting}. This approach leverages expert knowledge or human preference to assess the relative value of queried action pairs, using the preferred actions to guide policy updates. Compared to absolute scoring or full demonstration, action preference queries are often easier for humans to provide, as they rely on relative judgments rather than precise valuations.   This reduces the cognitive load on annotators while still yielding rich supervision signals.   In particular, action preference queries are especially valuable for identifying potentially high-value actions in regions where the current policy exhibits high uncertainty, enabling more targeted and efficient policy improvement. 

Despite its promising prospects, this direction remains relatively underexplored. The current method \cite{yang2023boosting} adopts a relatively simplistic strategy, selecting preferred actions solely based on their distance to the dataset and applying them as direct constraints during policy updates. Such strategy introduces two critical limitations. First, queried actions may lie far outside the dataset distribution, a phenomenon we refer to as \emph{query shift}. This shift can induce significant policy deviation, thereby compromising the reliability of offline data for accurate value estimation and limiting the generalizability of such methods to broader offline RL frameworks, particularly in value regularization algorithms. As illustrated in Figure \ref{fig:motivation}, when the queried action lies in out-of-distribution (OOD) region and its ground-truth Q-value is higher than that of the dataset action, the regularization term will elevate its Q-value, leading to inaccurate value estimation. This bias can further drive the policy toward the OOD region, resulting in unreliable updates and even catastrophic degradation in policy performance. Second, the informational potential of queried preferences is often underutilized, as they are typically treated as static supervision throughout training. However, as learning process progresses, the policy becomes more reliable as the distribution of the learned policy gradually aligns with the dataset distribution. Therefore, using static constraint terms in later stages of training can hinder policy learning. In this case, dynamically adjusting the regularization strength associated with preferred actions can help avoid ineffective constraints, ensure the effective utilization of preference information, and ultimately enhance policy performance.

To address the limitations of the existing action preference query method in offline RL, we propose novel techniques that focus on both how to query preferences and how to utilize them effectively. The core of our method lies in mitigating query shift and inefficient utilization by leveraging uncertainty estimation. Specifically, we employ the Morse neural network \cite{dherin2023morse,basu2023connection,zhuoutward,srinivasan2024offline} as an uncertainty estimator to quantify the OOD degree between the given actions and the offline dataset.
To query informative and helpful actions, we introduce the uncertainty-driven conservative query, which uses Morse-based proximity scores to filter and select candidate actions that are near the dataset distribution. 
This selective filtering avoids potentially significant policy deviation and value estimation errors and extends the applicability of the action preference query mechanism beyond policy constraint algorithms to broader offline RL paradigms such as value regularization. To exploit preference information effectively, we propose the uncertainty-aware adaptive regularization, which dynamically adjusts the data-level constraints based on the estimated uncertainty of the learned policy's actions. Actions that align closely with the dataset distribution are subjected to weaker regularization, preserving the potential benefit from Bellman updates, while stronger penalties are imposed on OOD actions to prevent unreliable update. This adaptive approach allows for more precise exploitation of query information and dynamically balances pessimistic constraints and optimistic Bellman updates. To validate the effectiveness of our approach in value regularization frameworks, we integrate it into CQL, a representative algorithm in this category, and conduct extensive experiments. Results on the D4RL benchmark demonstrate that our method achieves competitive performance across a diverse set of tasks. \

In summary, the primary contribution of our proposed approach can be outlined as follows:
\begin{itemize}
  \item We introduce the Morse Neural Network as an uncertainty estimator to quantify the OOD degree between the candidate actions and the offline dataset. Building on this estimator, we propose (i) the \textit{uncertainty-driven conservative query} mechanism to query informative and helpful actions, and (ii) the \textit{uncertainty-aware adaptive regularization} mechanism to adaptively adjust the constraint on preferred actions throughout training.
  
  \item We integrate the proposed lightweight approach into CQL, the applicability of the action preference query mechanism to value regularization algorithms. Extensive experiments on the D4RL benchmark demonstrate the competitive performance of our method.
\end{itemize}

\section{Preliminaries}
In this section, we present the necessary background on offline RL and the Morse neural network.
\paragraph{Offline RL} The environment in RL is typically modeled as a Markov Decision Process (MDP), defined by tuple $(\mathcal{S}, \mathcal{A}, P, r, \gamma)$, where $\mathcal{S}$ is the state space, $\mathcal{A}$ is the action space, $P(s'|s,a)$ denotes the transition probability from state $s$ to $s'$ under action $a$, $r(s,a)$ is the reward function, and $\gamma \in [0,1)$ is the discount factor. The agent interacts with the environment by selecting actions to maximize the expected cumulative reward. 

Offline RL, also known as batch RL, is a subfield of RL that focuses on learning policies solely from pre-collected datasets. To learn a reliable policy under distribution shift, existing methods commonly adopt the key principle of incorporating pessimism into policy learning, thereby constraining the learned policy to remain close to the dataset. In model-free RL, two essential components are the value function and the policy. To enforce pessimism, existing offline RL approaches typically fall into two categories: policy constraint methods and value regularization methods. 

Policy constraint methods aim to directly regularize the learned policy to remain close to the behavior policy. A unified optimization objective for this family of methods can be formulated as:
\begin{equation}
\underset{\pi}{\arg \max} \ \mathbb{E}_{s \sim \mathcal{D},\, a \sim \pi(\cdot|s)}\left[ Q(s, a) \right] - \lambda \cdot L_{\mathrm{PC}}\left(\pi \| \pi_\beta\right)
\end{equation}
where $\pi_\beta$ denotes the behavior policy, $L_{\mathrm{PC}}$ is a divergence measure (e.g., KL divergence), and $\lambda$ controls the strength of the constraint. Value functions in these methods are typically updated in an online manner.

In contrast, value regularization methods impose conservatism on the value function. The policy is updated using the learned Q-function in a standard online manner, while the value function is optimized with an additional regularization term. A general form of the objective for value regularization is:
\begin{equation}
\underset{Q}{\arg \min} \  \alpha \cdot \mathbb{E}_{s \sim \mathcal{D}, a\sim\rho(\cdot|s)}\left[Q(s,a)\right]  + L_{\mathrm{TD}} \notag
\end{equation}
\begin{equation}
L_{\mathrm{TD}} = \frac{1}{2} \mathbb{E}_{s, a \sim \mathcal{D}} \left[ \left( Q(s, a) - \hat{\mathcal{B}}^\pi \hat{Q}^k(s, a) \right)^2 \right]
\end{equation}
where $\alpha$ controls the regularization strength, $\rho$ represents a particular distribution, and $\hat{\mathcal{B}}^\pi \hat{Q}^k(s, a)$ denotes the experiential Bellman backup operator with respect to policy $\pi$.

\paragraph{Morse Neural Network} The Morse neural network \cite{dherin2023morse} is a recent technique designed to quantify uncertainty \cite{zhuoutward, srinivasan2024offline}. It can be used on top of a pre-trained network to bring distance-aware calibration with respect to the training data. For a given input $x$, the Morse network outputs a density score $M(x) \in [0, 1]$, where a value of 1 corresponds to the mode submanifolds of the data distribution, and the score decreases toward 0 as the input deviates from these high-density regions.

The threshold of rate decrease is controlled by the Morse Kernel, a positive
definite kernel $K$ defined as: \( K: \mathcal{Z} \times \mathcal{Z} \rightarrow [0, 1] \), where \( \mathcal{Z} = \mathbb{R}^k \) is a latent feature space. Such kernels satisfy \( K(z_1, z_2) = 1 \) if and only if \( z_1 = z_2 \), and typically take the exponential form \( K(z_1, z_2) = \exp(-D(z_1, z_2)) \), where \( D \) is a divergence measure. In this work, we adopt the Radial Basis Function (RBF) kernel as a specific instance:
\begin{equation}
    K_{\text{RBF}}(z_1, z_2) = \exp\left(-\frac{\lambda^2}{2} \| z_1 - z_2 \|^2 \right)
\end{equation}
where \( \lambda \) is a scale parameter. 

A Morse neural network is defined as a function $f_\phi: \mathcal{X} \rightarrow \mathcal{Z}$ combined with a Morse Kernel $K(z, t)$, where $t \subset \mathcal{Z}$ is a target chosen as a hyperparameter of the model. Consider a dataset $\mathcal{D} = \{x_1,...,x_n \}$ sampled from a data distribution with density $p(x)$. The Morse neural network is expressed as $M_\phi(x)=K(f_\phi(x),t)$ and trained by minimizing the KL divergence $D_{KL}\left(p(x) || M_\phi(x)\right)$ for unnormalized densities, as follows:
\begin{equation}
    \min_\phi \ \mathbb{E}_{x \sim p(x)}\left[\log\frac{p(x)}{M_\phi(x)} \right] + \int M_\phi(x) dx-\int p(x)dx
\end{equation}
which amounts to minimizing w.r.t. $\phi$ the following quantity:
\begin{equation} \label{morse_loss}
    \mathbb{E}_{x \sim p(x)}\left[-\log K(f_\phi(x),t)\right] + \mathbb{E}_{x \sim Uni}\left[ K(f_\phi(x),t)\right]
\end{equation}
It is easy to see that $M_\phi(x) \in [0, 1]$, and when $M_\phi(x) = 1$, $x$ corresponds to a mode that coincides with the level set of the submanifold of the Morse neural network.
Furthermore, $M_\phi(x)$ corresponds to the certainty of the sample $x$ being from the training dataset, so $1-M_\phi(x)$ is a measure of the epistemic uncertainty of $x$.

\smallskip

\section{Method}
As outlined in the introduction, both the design of the querying mechanism and the effective utilization of queried information remain underexplored, especially within value regularization methods. Given that the query shift problem primarily stems from selecting OOD actions, we adopt uncertainty estimation as the foundation of our approach. In Section \ref{PTUE}, we introduce the Morse neural network as an uncertainty estimator to quantify the proximity between candidate actions and the dataset distribution. Building on this, Section \ref{UDCQ} presents the \textit{uncertainty-driven conservative query} mechanism, aiming to ensure that the queried actions remain close to the dataset distribution, thereby preventing potential destabilization in subsequent learning updates. In Section \ref{UADR}, we further present \textit{uncertainty-aware adaptive regularization} to exploit preference information in an adaptive manner. This adaptive strategy modulates the degree of pessimism based on the estimated OOD degree of the current policy actions, striking a balance between conservative regularization and confidence in Bellman updates. Finally, in Section \ref{Algo}, we integrate these components into CQL, extending the action preference query mechanism to value regularization algorithms.

\subsection{Pre-trained Uncertainty Estimator}\label{PTUE}
Since querying OOD actions may result in inaccurate value estimation and policy deviation, we propose incorporating uncertainty estimation into the action preference query mechanism. To this end, we employ the Morse neural network, a recently proposed and effective method for quantifying uncertainty, as our uncertainty estimator. Our objective is to assess whether a given action is OOD. Specifically, we train the Morse neural network on the offline dataset $\mathcal{D}$, treating the input $x$ as state-action pairs $(s, a)$ and the target $t$ as actions $a$ in Eq. (\ref{morse_loss}). The corresponding optimization objective is formulated as follows:
\begin{equation}\label{pretrain}
\begin{split}
    \mathcal{L}_{\phi} = \mathbb{E}_{(s, a) \sim \mathcal{D}} \Big[ 
 -\log & K(f_\phi(s, a), a) \\
 + &\mathbb{E}_{a' \sim \mathcal{U}(\mathcal{A})} K(f_\phi(s, a'), a') 
\Big]
\end{split}
\end{equation}
where $K$ denotes the RBF kernel and $\mathcal{U}(\mathcal{A})$ denotes a uniform distribution over the action space.

After pre-training, the Morse neural network assigns higher scores to state-action pairs that lie within the dataset distribution and lower scores to out-of-distribution pairs, thereby effectively quantifying uncertainty. We refer to the scores $M_\phi(s,a)=K(f_\phi(s, a), a)$ as Morse scores, which represent the certainty that the state-action pairs originate from the offline dataset. Since this training procedure relies exclusively on the offline dataset, it does not incur additional computational overhead during the offline policy training phase and offers strong reusability across various algorithms or tasks.

\subsection{Uncertainty-Driven Conservative Query}\label{UDCQ}
As mentioned in Section \ref{introduction}, existing methods \cite{yang2023boosting} typically select queried actions solely based on a Euclidean distance criterion between actions sampled from the dataset and those generated by the policy, which can lead to \textit{query shift}. For value regularization methods, this shift can propagate through Bellman backups, resulting in unreliable value estimation. Therefore, it is crucial to avoid selecting OOD actions during the query process. Leveraging the pre-trained Morse neural network as an uncertainty estimator, we propose an uncertainty-driven conservative query mechanism that prioritizes in-distribution actions to enhance stability and reliability.

During the training of the Morse neural network, significant variations were observed in the mean and variance of the Morse scores across different datasets. This variability made it challenging to use a fixed threshold for OOD detection across different tasks. To address this issue, we propose a dataset-aware threshold based on the Morse score distribution of the dataset, defined as $\delta=\mu-n\sigma$, where $\mu$ and $\sigma$ represent the mean and standard deviation of the Morse score distribution of the dataset, respectively. We use $n$, referred to as the conservativeness coefficient, to control the conservativeness of the query process: a larger $n$ allows the queried actions to deviate further from the dataset distribution for potential performance gain, while a smaller $n$ restricts them to remain closer to the distribution for stable policy learning. 

In practice, we follow the approach of prior work \cite{yang2023boosting} by traversing the dataset to generate policy actions and then ranking them based on the Euclidean distance between each policy action $\hat{a}$ and the corresponding dataset action $a$. The candidate action set is then constructed by sorting these pairs in descending order of distance. To ensure that the queried actions are informative yet remain within the dataset distribution, we leverage the pre-trained Morse neural network to filter candidates using the precomputed threshold $\delta$. Specifically, only actions $\hat{a_i}$ satisfying $M_\phi(s, \hat{a_i})>\delta$ are retained for querying. After the uncertainty-driven conservative selection, we query preferences from an oracle over the candidate action set. In our setting, we use the Q-value function of a stronger policy, denoted as $Q^\star$, as the oracle. The preferred action is then determined by:
\begin{equation}
\tilde{a}=G(s, a, \hat{a}) = \underset{a' \in \{a, \hat{a}\}}{\arg\max} \ Q^*(s, a')
\label{UCQ}
\end{equation}
During policy learning, we select $N_q$ actions pairs with the highest divergence after the filtration by uncertainty at an interval of $T_q$ steps. In this work, we use an offline-to-online algorithm \cite{luo2024optimistic} to train the oracle, with details provided in Appendix B.

Building on the uncertainty-driven conservative query mechanism, the queried actions remain close to the dataset distribution, effectively mitigating the query shift. This helps ensure stable policy learning, especially in value regularization methods. The near-distribution preferred actions can be reliably evaluated by the Q-value function trained on the offline data, reducing the risk of OOD behaviors and enhancing the effectiveness of Bellman updates.

\subsection{Uncertainty-Aware Adaptive Regularization}\label{UADR}
Previous work directly models the preference between policy actions and those from the dataset in the manner of semi-supervised learning, replacing the dataset actions in the regularization term with all preferred actions \cite{yang2023boosting}. This approach introduces additional computational cost. Moreover, purely data-driven semi-supervised methods may struggle to capture complex preference relationships, especially in continuous spaces. To exploit queried information in a more reliable and effective way, we focus on the necessity of regularization rather than attempting to generalize to potentially unreliable preferred actions. We propose uncertainty-aware adaptive regularization to avoid ineffective constraints and benefit from the Bellman updates. Specifically, we replace the fix regularization with data-level regularization determined by the OOD degree of the policy action, balancing the confidence in Bellman updates with the pessimism induced by over-regularization. We implement this mechanism by replacing the fixed regularization coefficient $\alpha$ with the data-dependent regularization coefficients $\alpha(s,a)$, defined based on the Morse scores as:
\begin{equation}\label{linear alpha}
\alpha(s,a) = (1 - M_\phi(s,a)) \cdot \alpha_0
\end{equation}
where $\alpha_0$ is the pre-defined value representing the maximum regularization strength.
When the OOD degree of a policy action is low, constraints on the preferred action are unnecessary, as the policy action already resides in a reliable region. In such cases, the proposed adaptive regularization enables the policy to fully benefit from Bellman updates. Conversely, when the policy action is likely OOD and thus unreliable, stronger regularization on reliable preferred actions guides the policy update toward the dataset distribution, promoting stability and safety in policy learning.

\begin{figure}[t]
\begin{center}
\centerline{\includegraphics[width=\columnwidth]{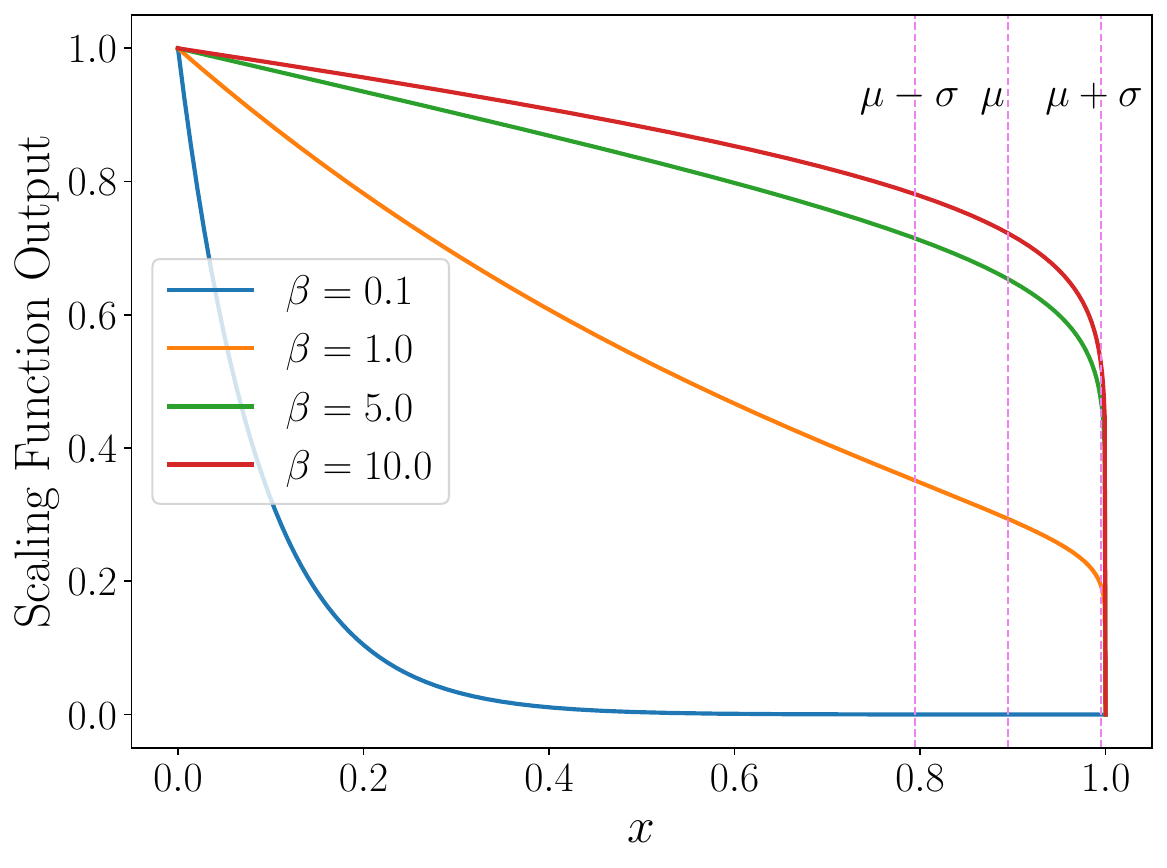}}
\caption{The variation of the nonlinear scaling function in Eq. (\ref{scaling term}) under different values of $\beta$}
\label{fig:scaling_function}
\end{center}
\end{figure}

However, we observe that most Morse scores tend to be relatively high during training, resulting in consistently low $\alpha$ values. This leads to overly weak regularization and insufficient discriminative ability. To achieve more discriminative regularization, we incorporate the mean and standard deviation of the Morse score distribution of the offline dataset and reformulate the linear relationship in Eq. (\ref{linear alpha}) into a nonlinear form, as follows:
\begin{equation}\label{scaling term}
\mathcal{S}_\beta(s,a) =  \left(1-M_\phi(s,a)\right)^{\sigma} \cdot e^{ M_\phi(s,a)/(\beta\cdot\mu)}
\end{equation}
\begin{equation}\label{adaptive coef}
    \alpha(s,a)=  \mathcal{S}_\beta(s,a)\cdot \alpha_0
\end{equation}
where $\beta$ is the hyperparameter that controls the slope of the coefficient function. The first term in Eq. (\ref{scaling term}) preserves the original value range, while the second term adjusts the regularization strength for different OOD degrees. As shown in Figure \ref{fig:scaling_function}, a larger $\beta$ leads to a stronger constraint when the Morse score falls below $\mu - \sigma$, while a smaller $\beta$ imposes such a constraint only when the Morse score is significantly small. In all cases, when the Morse score approaches 1, indicating that the action lies within the dataset distribution, the regularization coefficient approaches 0, effectively avoiding unnecessary constraints.

Through uncertainty-aware adaptive regularization with nonlinear scaling, we have achieved discriminative constraints on different data, avoiding excessive regularization on preferred actions while leveraging the potential benefits of Bellman updates. Since preference queries are conducted at intervals during training, such differentiated constraints are essential. They prevent ineffective constraints from outdated preferences on the current well-learned policy and enable the exploration of better actions by capitalizing on the optimistic nature of Bellman updates when policy actions fall within the dataset distribution.

The proposed adaptive regularization mechanism and conservative querying strategy are inherently complementary. When the policy action exhibits a low degree of OOD, meaning that it falls within the support of the offline dataset, both the regularization strength and the query probability remain low. In this case, policy updates are primarily guided by Bellman backups, enabling optimistic updates that approximate online learning. In contrast, when the policy action is highly OOD, the regularization becomes stronger, and the update is driven more by the queried preference signal. If the action deviates significantly from the dataset distribution, it is excluded from the set of query candidates, and the update is constrained by actions observed in the dataset. When the policy action is moderately OOD, that is, situated near the boundary of the data distribution, querying its relative preference acts as a form of localized exploration within the data manifold, effectively expanding the trusted region for learning.

\subsection{Extend to Value Regularization Methods} \label{Algo}
Our proposed approach is algorithm-agnostic, enabling seamless integration with a broad class of offline RL algorithms, particularly those based on value regularization. To empirically validate our approach, we integrate it into Conservative Q-Learning (CQL) \cite{kumar2020conservative}, a representative value regularization method. CQL mitigates the overestimation of OOD actions by pessimistically pushing down their estimated Q-values, encouraging the learned policy to remain close to the dataset distribution. Specifically, CQL introduces an additional regularization term in the value function objective, which simultaneously raises Q-values for the dataset actions while penalizing those of the current policy. The optimization objective of the value function in CQL is formulated as follow:
 \begin{equation}
 \begin{split}
\min_Q\ \alpha \cdot \mathbb{E}_{s \sim \mathcal{D}} 
\left[ 
\log \sum_{a} \exp(Q(s, a)) 
- \mathbb{E}_{a \sim D} \left(Q\left(s, a\right) \right) 
\right]  + \\
\frac{1}{2} \mathbb{E}_{(s, a) \sim \mathcal{D}} 
\left[ 
\left( Q(s, a) - \hat{\mathcal{B}}^\pi \hat{Q}^k(s, a) \right)^2 
\right]
 \end{split}
 \end{equation}

To facilitate the computation for the Morse score of the current policy actions and reduce computational overhead, we integrate the proposed method into the original version of CQL and name it Conservative Query Q-Learning (CQ2L), resulting in the following optimization objective:
\begin{equation}\label{pi_loss}
\max_\pi  \mathbb{E}_{s \sim \mathcal{D},a\sim \pi(\cdot|s)} \left[Q(s,a)- \omega \log \pi(a|s) \right]
\end{equation}
\begin{equation}\label{q_loss}
\begin{split}
\min_Q  \mathbb{E}_{s \sim \mathcal{D},a\sim \pi(\cdot|s)} \left[\alpha(s,a)  \left[Q(s,a)-  Q\left(s, \tilde{a} \right) \right]  \right] \\
 + \frac{1}{2} \mathbb{E}_{(s, a) \sim \mathcal{D}}
\left[ \left( Q(s, a) - \hat{\mathcal{B}}^\pi \hat{Q}^k(s, a) \right)^2 \right]
\end{split}
\end{equation}
where $\tilde{a}$ denotes the preferred action. 
In practice, the preferred actions are initially set as dataset actions, and they are later updated with the corresponding preferred actions after the preference queries are conducted.
The pseudo code is presented in Algorithm \ref{alg:conservative_q_learning}, with components specific to our approach highlighted in red.

Compared to previous work, the advantage of our method lies in its ability to alleviate query shift through the uncertainty-driven conservative query mechanism, preventing incorrect value overestimation. Additionally, by employing uncertainty-aware adaptive regularization, our method harnesses the potential benefits of Bellman updates, which is crucial for enhancing the policy performance of value regularization methods.

\begin{algorithm}[t]
\caption{Conservative Query Q-Learning (CQ2L)}\label{alg:conservative_q_learning}
\begin{algorithmic}[1]
\Statex \textbf{Input:} Offline dataset $\mathcal{D}$, oracle $Q^*$, Morse neural network $M_\phi$, conservativeness coefficient $n$, query interval $T_q$, number of preference queries per query step $N_p$, empty replay buffer $R$
\State Initialize Q-function $Q_{\theta}$, policy $\pi_{\psi}$ with random parameters
\State Initialize preferred actions $\tilde{a}$ with the dataset actions $a$ from $\mathcal{D}$
\State Initialize replay buffer $R \leftarrow \{\mathcal{D}, \tilde{a}\}$

\Statex \textit{\textcolor{red}{\# Pre-training Stage}}
\State Pre-train $M_\phi$ on $\mathcal{D}$ by Eq. (\ref{pretrain}) 
\State Compute Morse scores on $\mathcal{D}$ to obtain mean $\mu$ and standard deviation $\sigma$, and set threshold $\delta = \mu - n \sigma$

\Statex \textit{\# Offline Training Stage}
\For{iteration $i = 1, 2, \ldots$}
    \State Sample mini-batch $\{s, a, r, s', \tilde{a}\} \sim R$
    \State Update policy $\pi_\psi$ by Eq. (\ref{pi_loss})
    \State \textcolor{red}{Update Q-function $Q_\theta$ by Eq. (\ref{q_loss})}
    \If{$i \bmod T_q = 0$}
        \State Compute Euclidean distances between policy actions $\hat{a} = \pi_\psi(s)$ and dataset actions $a$ for all $s$ in $R$
        \State Sort action pairs $(\hat{a}, a)$ by descending distance
        \State \textcolor{red}{Select top-$N_p$ pairs with $M_\phi(s, \hat{a}) > \delta$ for querying}
        \State Update the corresponding $\tilde{a}$ in $R$ based on queried results
    \EndIf
\EndFor
\end{algorithmic}
\end{algorithm}

\begin{table*}[ht]
\caption{\textbf{Offline} scheme performance comparison on average normalized D4RL score over the final 10 evaluations and 5 random seed.}
\vspace{0.3cm}
\centering
\begin{tabular}{l c c c c c c c c}
\hline
\textbf{Dataset} & \textbf{AWAC} & \textbf{IQL} & \textbf{SPOT} & \textbf{Cal-QL} & \textbf{TD3+BC} & \textbf{CQL} & \textbf{CPI} & \textbf{CQ2L} \\
\hline
halfcheetah-medium-v2        & 49.8 ± 0.3  & 48.1 ± 0.3  & 57.6 ± 0.5  & 47.8 ± 0.2   & 48.3 ± 0.3   & 47.0 ± 0.2   & 64.4 ± 1.3   & \textbf{67.8 ± 0.6} \\
hopper-medium-v2             & 68.6 ± 11.2 & 66.7 ± 4.4  & 71.4 ± 16.0 & 64.7 ± 3.4  & 58.7 ± 5.9   & 65.3 ± 3.3   & \textbf{98.5 ± 3.0}   & 90.2 ± 6.1 \\
walker2d-medium-v2           & 85.1 ± 0.5  & 74.8 ± 1.8  & 69.6 ± 0.6  & 84.3 ± 0.9  & 82.3 ± 1.4   & 81.2 ± 1.4   & 85.8 ± 0.8   & \textbf{87.4 ± 1.3} \\
halfcheetah-medium-replay-v2 & 45.4 ± 0.6  & 44.5 ± 0.3  & 52.3 ± 1.3  & 46.2 ± 0.3  & 44.4 ± 0.6   & 45.2 ± 0.5   & 54.6 ± 1.3   & \textbf{61.5 ± 3.1} \\
hopper-medium-replay-v2      & 97.8 ± 1.4  & 89.6 ± 11.9 & 87.1 ± 14.2 & 93.4 ± 6.6  & 66.4 ± 25.0  & 92.4 ± 9.5   & 101.7 ± 1.6   & \textbf{101.8 ± 0.6} \\
walker2d-medium-replay-v2    & 73.2 ± 8.4  & 80.6 ± 5.8  & 88.9 ± 1.5  & 84.7 ± 1.4  & 81.6 ± 3.1   & 80.9 ± 2.1   & 91.8 ± 2.9   & \textbf{93.0 ± 1.2} \\
halfcheetah-medium-expert-v2 & 95.3 ± 0.9  & 91.8 ± 2.1  & 92.7 ± 2.8  & 52.7 ± 5.4  & 92.9 ± 4.6   & 93.7 ± 2.9   & 94.7 ± 1.1   & \textbf{95.4 ± 2.5} \\
hopper-medium-expert-v2      & 110.6 ± 1.3 & 106.3 ± 7.4 & 102.1 ± 14.8 & 107.6 ± 2.4 & 101.4 ± 4.9  & 98.3 ± 9.1   & 106.4 ± 4.3   & \textbf{111.3 ± 1.4} \\
walker2d-medium-expert-v2    & 89.7 ± 39.4 & 111.9 ± 1.0 & 110.3 ± 0.4 & 109.0 ± 0.3 & 110.3 ± 0.8  & 109.2 ± 0.2  & 110.9 ± 0.4   & \textbf{114.6 ± 3.8} \\
\hline
\textbf{Average MuJoCo}      & 79.28       & 79.37       & 81.33  & 76.71  & 76.3  & 79.1   & 89.9   & \textbf{91.4} \\
\hline
antmaze-umaze-v2             & 63.5 ± 19.2 & 74.8 ± 5.8  & 88.8 ± 3.3  & 74.8 ± 1.8   & 88.6 ± 4.6   & 90.0 ± 4.5   & \textbf{98.8 ± 1.1}   & 97.4 ± 1.5 \\
antmaze-umaze-diverse-v2     & 57.8 ± 8.0  & 52.2 ± 6.4  & 41.5 ± 5.3  & 16.2 ± 20.1 & 43.2 ± 18.8  & 27.8 ± 10.5  & \textbf{88.6 ± 5.7}   & 62.8 ± 9.4 \\
antmaze-medium-play-v2       & 0.0 ± 0.0   & 63.8 ± 4.8  & 63.0 ± 13.9 & 69.5 ± 8.0   & 0.0 ± 0.0   & 68.5 ± 3.8   & 82.4 ± 5.8   & \textbf{83.6 ± 6.4} \\
antmaze-medium-diverse-v2    & 0.0 ± 0.0   & 61.2 ± 2.5  & 67.0 ± 21.5 & 64.0 ± 5.6  & 0.0 ± 0.0   & 64.0 ± 6.7   & \textbf{80.4 ± 8.9}   & 72.2 ± 11.8 \\
antmaze-large-play-v2        & 0.0 ± 0.0   & 37.8 ± 3.8  & 34.0 ± 7.1  & 41.8 ± 4.5   & 0.0 ± 0.0   & 23.5 ± 5.1   & 20.6 ± 16.3   & \textbf{65.4 ± 5.5} \\
antmaze-large-diverse-v2     & 0.0 ± 0.0   & 20.8 ± 6.1  & 36.2 ± 11.6 & 32.8 ± 9.3   & 0.0 ± 0.0    & 22.5 ± 14.0  & 45.2 ± 6.9   & \textbf{73.2 ± 3.4} \\
\hline
\textbf{Average Antmaze}     & 20.22       & 51.77       & 55.08  & 49.85  & 22.0   & 49.0   & 69.3   & \textbf{75.8} \\
\hline
\end{tabular}
\label{tab:offline}
\end{table*}

\begin{table*}[ht]
\centering
\caption{\textbf{O2O} scheme performance comparison on average normalized D4RL score over the final 10 evaluations and 5 random seed. Our method achieves superior or comparable performance with only 90,000 preference queries, while OAP conducted 100,000 queries and other methods interacted with the environment by an additional 250,000 steps. The second-best average performance is highlighted with the underline}
\vspace{0.36cm}
\begin{tabular}{l c c c c c c c}
\hline
\textbf{Dataset} & \textbf{AWAC} & \textbf{IQL} & \textbf{SPOT} & \textbf{TD3+BC} & \textbf{CQL} & \textbf{OAP}  & \textbf{CQ2L} \\
\hline
halfcheetah-medium-v2        & 56.7 ± 1.4 & 49.7 ± 0.2 & 58.6 ± 0.8 & 52.5 ± 0.5 & 48.0 ± 0.2 & 56.4 ± 4.3 & \textbf{67.8 ± 0.6} \\
hopper-medium-v2             & 98.7 ± 3.6 & 75.2 ± 4.5 & \textbf{99.9 ± 0.3} & 63.7 ± 7.5 & 63.8 ± 3.6 & 82.0 ± 6.6 & 90.2 ± 6.1 \\
walker2d-medium-v2           & 87.1 ± 0.6 & 80.8 ± 6.8 & 82.5 ± 1.7 & 86.6 ± 0.8 & 82.8 ± 0.5 & 85.6 ± 1.2 & \textbf{87.4 ± 1.3} \\
halfcheetah-medium-replay-v2 & 49.3 ± 0.5 & 45.2 ± 1.2 & 57.6 ± 1.0 & 49.3 ± 2.6 & 49.4 ± 0.1 & 53.4 ± 1.9 & \textbf{61.5 ± 3.1} \\
hopper-medium-replay-v2      & 100.0 ± 1.0 & 91.1 ± 12.9 & 97.3 ± 2.1 & 97.0 ± 1.2 & 101.3 ± 0.3 & 98.5 ± 2.5 & \textbf{101.8 ± 0.6} \\
walker2d-medium-replay-v2    & \textbf{94.2 ± 3.8} & 89.2 ± 6.5 & 86.4 ± 3.4 & 89.9 ± 3.3 & 87.9 ± 2.3 & 84.3 ± 2.7 & 93.0 ± 1.2 \\
halfcheetah-medium-expert-v2 & 95.1 ± 0.6 & 92.4 ± 1.9 & 91.9 ± 1.1 & 93.2 ± 1.0 & \textbf{95.7 ± 1.0} & 83.4 ± 5.3 & 95.4 ± 2.5 \\
hopper-medium-expert-v2      & \textbf{112.1 ± 0.7} & 109.6 ± 4.0 & 106.5 ± 0.5 & 99.8 ± 13.2 & 110.8 ± 0.2 & 85.9 ± 6.6 & 111.3 ± 1.4 \\
walker2d-medium-expert-v2    & 113.5 ± 2.4 & 115.0 ± 0.4 & 110.6 ± 0.4 & \textbf{115.8 ± 0.6} & 109.8 ± 0.2 & 111.1 ± 0.6 & 114.6 ± 3.8 \\
\hline
\textbf{Average MuJoCo}      & 84.1 & 83.2 & \underline{88.1} & 83.08 & 82.39 & 82.3 & \textbf{91.4} \\
\hline
antmaze-umaze-v2             & 92.2 ± 4.0 & 83.0 ± 4.1 & \textbf{99.8 ± 0.4} & 72.8 ± 23.8 & 95.2 ± 1.3 & 90.4 ± 5.2 & 97.4 ± 1.5 \\
antmaze-umaze-diverse-v2     & 3.0 ± 4.6 & 38.2 ± 17.6 & 56.8 ± 18.8 & 39.8 ± 13.0 & 59.2 ± 28.4 & \textbf{75.0 ± 19.0} & 62.8 ± 9.4 \\
antmaze-medium-play-v2       & 0.0 ± 0.0 & 78.8 ± 6.4 & \textbf{92.5 ± 2.5} & 0.0 ± 0.0 & 77.0 ± 4.4 & 62.0 ± 10.0 & 83.6 ± 6.4 \\
antmaze-medium-diverse-v2    & 0.0 ± 0.0 & 80.2 ± 5.3 & \textbf{87.0 ± 7.0} & 0.2 ± 0.4 & 84.0 ± 3.0 & 54.5 ± 23.3 & 72.2 ± 11.8 \\
antmaze-large-play-v2        & 0.0 ± 0.0 & 42.8 ± 3.9 & 60.0 ± 9.6 & 0.0 ± 0.0 & 51.8 ± 12.1 & 0.0 ± 0.0 & \textbf{65.4 ± 5.5} \\
antmaze-large-diverse-v2     & 0.0 ± 0.0 & 40.2 ± 7.9 & 63.0 ± 4.2 & 0.0 ± 0.0 & 38.2 ± 14.0 & 9.4 ± 8.4 & \textbf{73.2 ± 3.4} \\
\hline
\textbf{Average Antmaze}     & 15.87 & 60.5 & \textbf{76.35} & 18.8 & 67.56 & 48.6 & \underline{75.8} \\
\hline
\end{tabular}
\label{tab:o2o}
\end{table*}

\setlength{\abovecaptionskip}{-8pt}

\begin{figure*}[ht]
\centerline{\includegraphics[width=\textwidth]{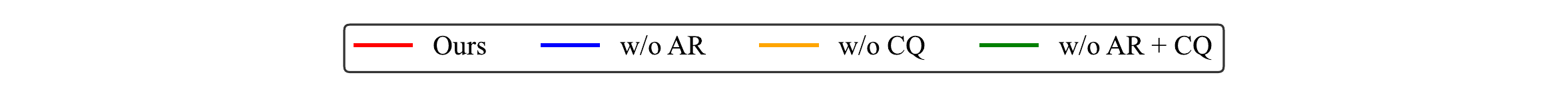}}
    \begin{center}
    \subfigure[]{
        \label{a}
        \includegraphics[width=0.23\linewidth]{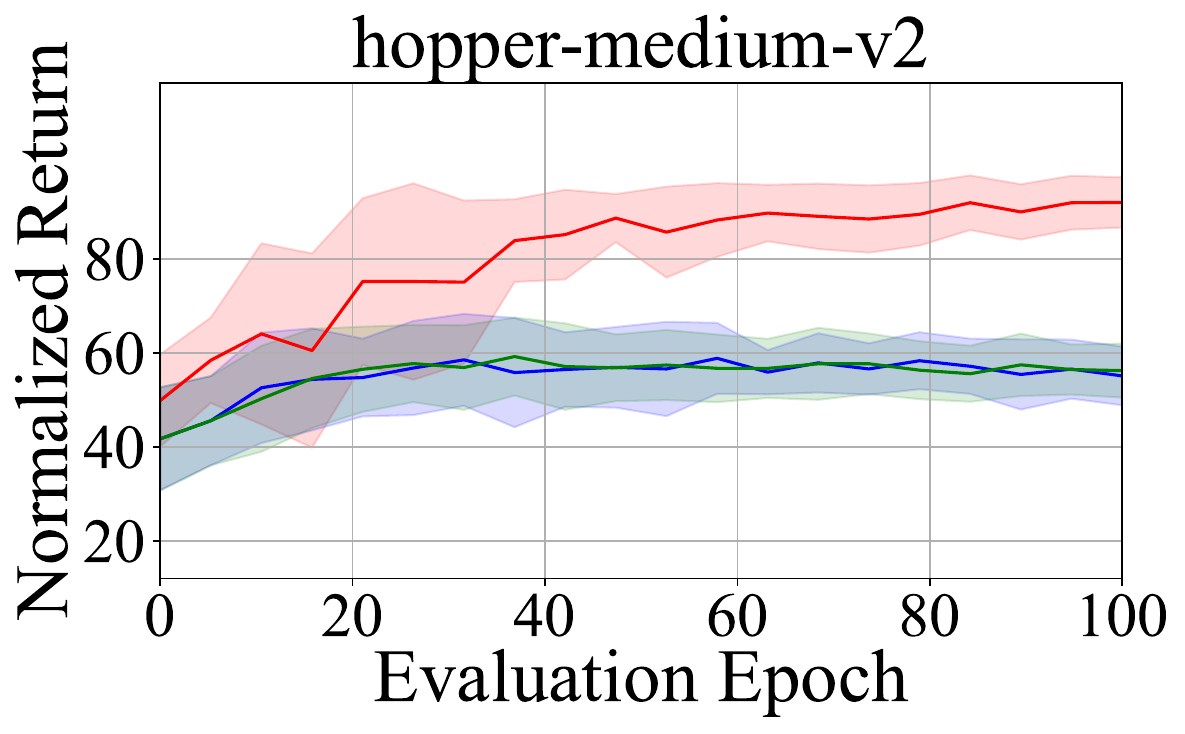}}
    \subfigure[]{
        \label{b}
        \includegraphics[width=0.23\linewidth]{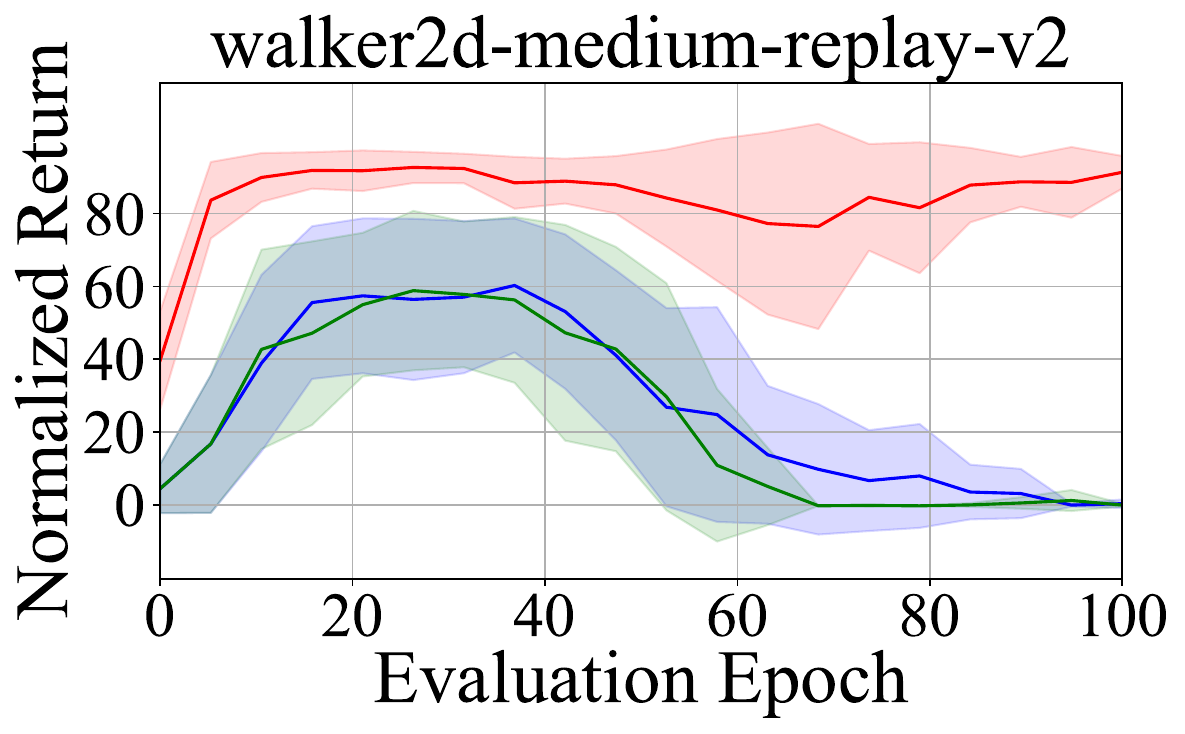}}
    \subfigure[]{
        \label{c}
        \includegraphics[width=0.23\linewidth]{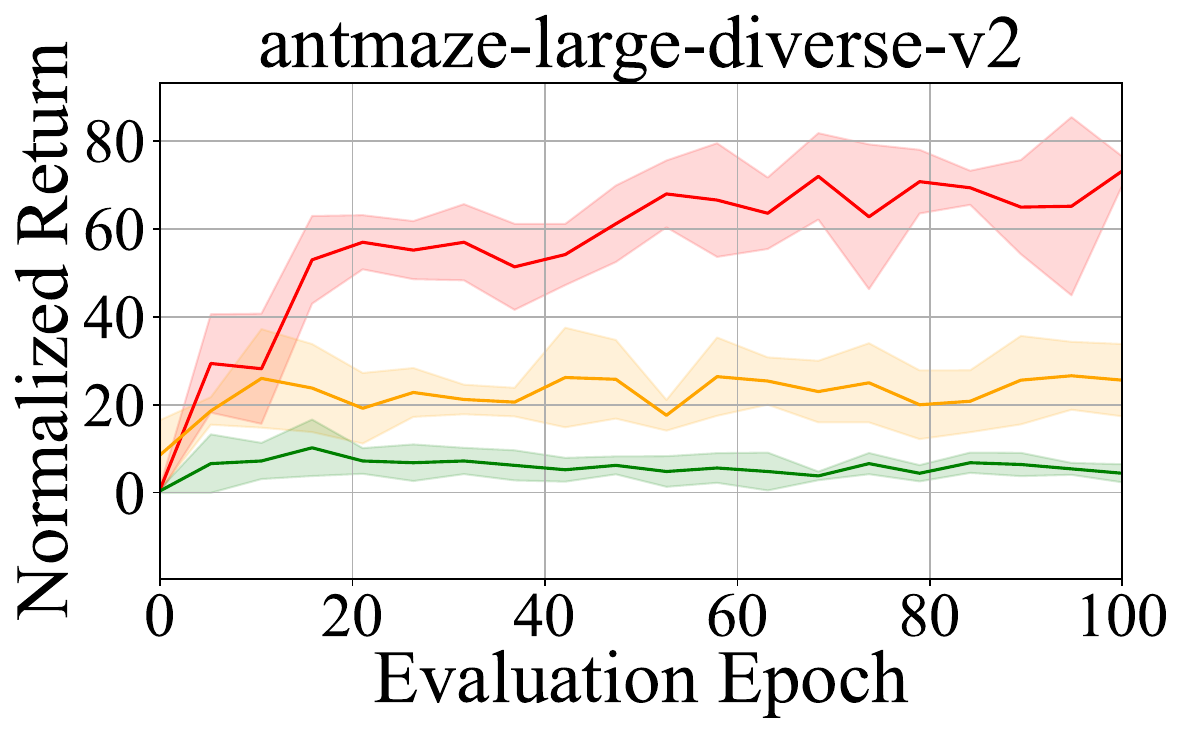}}
    \subfigure[]{
        \label{d}
        \includegraphics[width=0.23\linewidth]{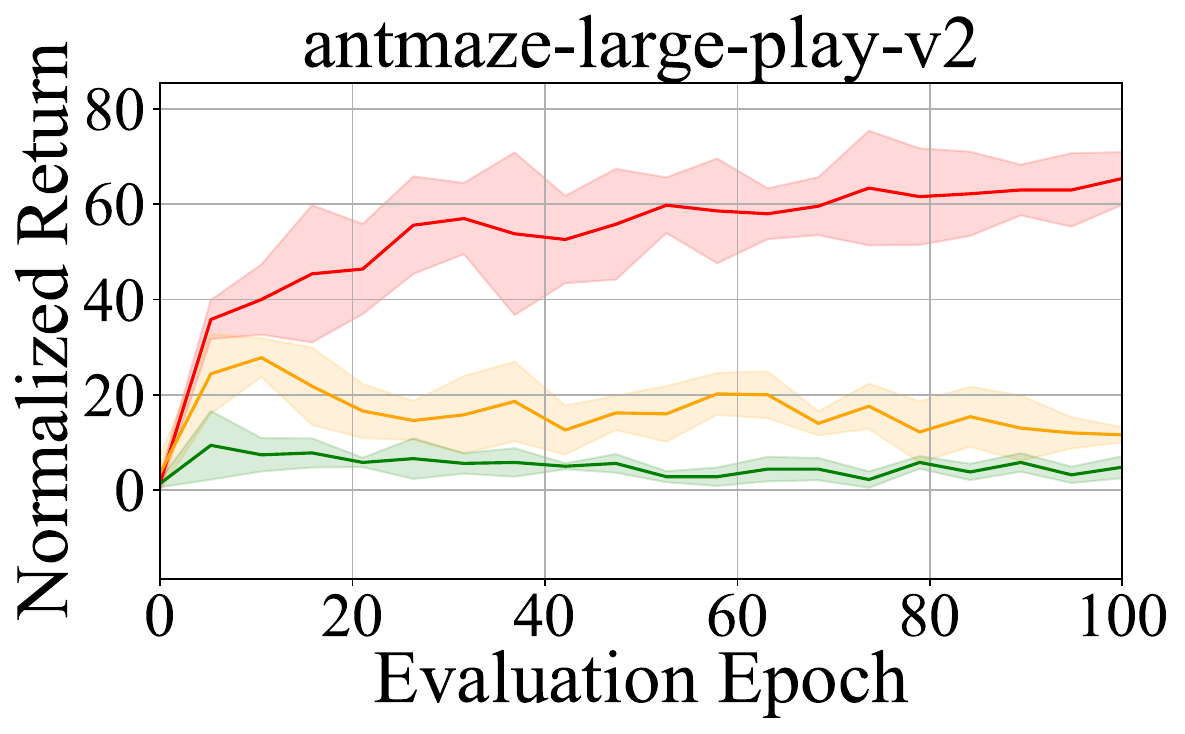}} 
    \end{center}

\caption{Ablation study across four tasks. figure \ref{a} and \ref{b} correspond to MuJoCo tasks with dense rewards and low-dimensional observations, while figure \ref{c} and \ref{d} correspond to Antmaze tasks with sparse reward and high-dimensional observations, which are much more challenging. }
\vspace{0.2cm}
\label{fig:ablation}
\end{figure*}

\section{Experiment}
In this section, we conducted extensive experiments to validate the effectiveness of CQ2L on the D4RL\cite{fu2020d4rl} benchmark. In Section \ref{setting}, we introduce the task domains and comparison schemes. Then, in Section \ref{comparison}, we demonstrate the superior performance of our method, especially against the baseline algorithm CQL and the previous action preference query method. Last but not least, to validate the contributions of individual components, we perform ablation studies in Section \ref{ablation} and provide empirical analysis.

\subsection{Experiment Setting}\label{setting}
\paragraph{Domains} We evaluate our approach on two widely-used task domains from the D4RL \cite{fu2020d4rl} benchmark: MuJoCo locomotion and AntMaze navigation. MuJoCo locomotion domain includes standard continuous control environments. In this paper, we select HalfCheetah, Hopper, and Walker2d, each with three dataset types: medium, medium-replay, and medium-expert. These tasks assess an algorithm’s ability to learn effective policies from offline trajectories of diverse sub-optimality. The AntMaze navigation domain poses a much greater challenge due to long-horizon planning, sparse rewards, and the need for precise locomotion. It involves guiding an 8-DoF ant-like quadruped robot navigate through a maze and reach a fixed goal location. There are three Antmaze environments with increasing complexity: umaze, medium, and large, each with two variants: play and diverse, introducing additional difficulty through increased environment complexity and sparse rewards.

\paragraph{Schemes} Consider that 
although our method is applied to offline training, the action preference query mechanism introduces additional information. In order to verify the effectiveness of the proposed method, we compared it with the baseline methods of two schemes: the offline setting and offline-to-online setting. Additionally, we include a comparison with OAP \cite{yang2023boosting} in O2O scheme, the only existing offline RL method that incorporates the action preference query.
Since OAP does not provide open-source code, we adopt the performance results reported in the original paper \cite{yang2023boosting}. For the remaining methods, we used the official code or implementations from open-source libraries to report the mean and standard deviation across five random seeds.
The offline training is conducted for 1,000,000 steps, and the online fine-tuning is performed for 250,000 steps.
In the implementation of OAP, a total of 100,000 action preference queries were conducted.
For our proposed CQ2L, the total preference query budget is set to 90,000, which is substantially fewer than the interactions used in online fine-tuning. Additional implementation details are provided in Appendix B.

\subsection{Performance Comparison}\label{comparison}
Table \ref{tab:offline} presents the comparisons of the average normalized scores on the D4RL benchmark across nine MuJoCo tasks and Antmaze tasks, evaluated over 5 random seeds. First, we observe that conventional offline RL methods, including both policy constraint and value regularization approaches, perform significantly worse than our proposed method. This performance gap highlights the advantage of incorporating the action preference query, which purposefully introduces beneficial prior knowledge (oracle evaluations) during training. 

Compared to online fine-tuning methods, our approach remains highly competitive. As shown in Table \ref{tab:o2o}, it outperforms most baselines that undergo 250,000 online interaction steps, using only 90,000 action preference queries.
Notably, our method achieves superior or comparable performance to the fine-tuned baseline CQL across most tasks. In contrast, OAP yields only marginal gains over its baseline method TD3+BC, and performs noticeably worse on the MuJoCo tasks.
These results highlight the clear advantage of our approach over OAP. 
Specifically, our method achieves substantially better performance than OAP on all tasks except \textit{antmaze-umaze-diverse-v2}, indicating that the proposed uncertainty-driven conservative query is more beneficial for policy learning and the uncertainty-aware adaptive regularization facilitates more effective exploitation of queried information and more reliable Bellman updates. 

We also applied OAP to value regularization methods built upon CQL. However, as shown in Appendix A, the performance degrades substantially, highlighting the limitations of extending OAP in this context. In contrast, our method successfully integrates the action preference query mechanism with value regularization, yielding consistently strong results.

\subsection{Ablation Study}\label{ablation}
We conduct ablation studies across four different tasks to assess the contributions of two key components in our method: Uncertainty-Driven Conservative Query (denoted as CQ) and Uncertainty-Aware Adaptive Regularization (denoted as AR). 

Figures \ref{a} and \ref{b} present the ablation results on the MuJoCo tasks. Owing to the dense reward signals in these environments, the adaptive regularization mechanism can effectively guide policy updates through more reliable Bellman backups. Furthermore, due to low-dimensional observations, the offline datasets offer high coverage, reducing the influence of whether a conservative strategy is applied in the action preference queries. 

For the more challenging tasks in the Antmaze domain, high-dimensional observations make it difficult for offline datasets to adequately cover the state and action spaces, while sparse reward signals hinder the ability to achieve reliable Bellman updates. Therefore, as shown in Figures \ref{c} and \ref{d}, conservative queries help ensure that the preferred actions remain close to the dataset distribution, avoiding the overestimation of infeasible OOD actions. Moreover, in essence, this resembles intra-distribution exploration, where the queried preferred actions help bridge previously disconnected states in the dataset, effectively enabling a form of trajectory stitching. This mechanism facilitates the exploitation of potentially helpful information from suboptimal trajectories, which is particularly beneficial for long-horizon planning tasks.

\section{Related Work}
In this section, we provide an overview of offline RL and preference-based RL by reviewing relevant literature.
\paragraph{Offline RL} 
As previously discussed, offline RL addresses the problem of distributional shift by avoiding extrapolation to unseen actions that lie outside the support of the dataset. Given an offline dataset collected by an unknown behavioral policy $\pi_\beta$, several methods \cite{fujimoto2018off,wu2019behavior,kostrikov2021offline,matsushima2020deployment} directly estimate $\pi_\beta$ and convert the unconstrained policy optimization objective into a constrained one \cite{prudencio2023survey} by enforcing a distributional constraint, which encourages the learned policy to match the behavioral policy. However, a key limitation of such direct policy constraint methods is their dependence on accurate estimation of $\pi_\beta$, since the behavior policy can be difficult to accurately recover. Some works avoid the explicit estimation of $\pi_\beta$ and enforce support-matching constraints \cite{wu2019behavior,fujimoto2021minimalist,peng2019awr,nair2021awac}, also referred to as implicit policy constraints. Value Regularization \cite{kumar2020conservative,nachum2019algaedice,kostrikov2021offline,kostrikov2022offline, yu2021combo} is another way to learn the policy without relying on $\pi_\beta$. Compared to policy constraint, value regularization methods do not restrain the policy directly. Instead, they penalize the values of OOD actions with regularization terms, and update the value function in a conservative manner.

\paragraph{Uncertainty estimation in offline RL} Some offline RL methods achieve conservative learning through uncertainty estimation. The key is to construct an uncertainty estimator. One typical approach \cite{agarwal2020optimistic, an2021uncertainty, nikulin2022q} is to train an ensemble of Q-functions, and use the average or minimum Q-values for conservative estimation. In this way, the value overestimation is mitigated, and the policy tends to output actions constrained within the dataset distribution. Another approach is to train an ensemble of transition models \cite{kidambi2020morel, yu2020mopo, yu2021combo} instead of estimating Q-function, primarily used in Model-based RL. They quantify the uncertainty by calculating the standard deviation of the predictions on the next state and add it as the penalty term to the reward function. By avoiding transitions to uncertain regions, these methods achieve conservative updates supported by the distribution of the dataset.

\paragraph{Preference-base RL} Preference-based RL (PBRL) is a promising approach to alleviate the necessity of precise reward design \cite{busa-fekete2014survey,akrour2011preference}. According to the query target, PBRL can be classified as the trajectory preference query and the state preference query. The former \cite{busa2013preference,wilson2012bayesian, christiano2017deep} queries the better trajectory in the given trajectory pair, and is the most general form of PBRL, as all preferences can be assigned to trajectory preferences when we admit trajectories with only a single element \cite{wirth2017survey}. The latter paradigm \cite{runarsson2012imitating,wirth2014learning,zucker2010optimization} identifies the state in a given state pair where there exists an action that is better than all the available actions in another state. Compared with the two typical paradigms, previous work \cite{furnkranz2012preference} on the action preference query is limited, which queries the better action in the given action pair under the same state. Recently, \citet{yang2023boosting} proposed OAP, introducing this paradigm to offline RL. OAP queries the preference between the policy actions and the dataset actions, and replaces the target constraint actions with the preferred actions based on TD3+BC \cite{fujimoto2021minimalist}. In addition, semi-supervised learning is introduced to generalize preference detection to uncompared regions, guiding the policy update towards the potentially beneficial direction and breaking the performance bottleneck induced by the limited distribution of the dataset. However, a key shortcoming of the query mechanism lies in their neglect for the update dynamics related to Bellman backups. This omission can result in unstable Q-function updates when preference actions are directly integrated into the critic learning process. To address this issue, our conservative query mechanism achieves stable value updates by selecting preferred actions near the dataset, and our adaptive regularization enables the effective exploitation of the potential benefits from more reliable Bellman updates. Furthermore, our method extends preference queries from policy optimization to value function learning, offering a novel insight for future research in this area.

\section{Conclusion}

To further explore the potential of the action preference query mechanism and extend its applicability to value regularization methods, we propose a framework that incorporates conservative querying and adaptive regularization under uncertainty estimation. Specifically, we employ a Morse neural network as an uncertainty estimator, pre-trained on the offline dataset to achieve the ability to quantify the OOD degree of given state-action pairs. To mitigate the risk of misleading policy updates caused by highly OOD actions, we introduce an uncertainty-driven conservative query mechanism that filters queried actions via the Morse network, ensuring that the preferred actions lie close to the data distribution and avoiding erroneous overestimation of OOD actions. Additionally, to prevent overly pessimistic constraints on preferred actions and to balance the influence of preference supervision with Bellman updates, we propose an uncertainty-aware adaptive regularization mechanism. This mechanism dynamically adjusts the regularization strength based on the OOD level of the current policy actions, enabling data-level discriminative constraints and allowing the agent to benefit from the strengths of Bellman backups. We integrate the proposed components into the representative value regularization method CQL and conduct extensive evaluations across diverse tasks on the D4RL benchmark. Experimental results show that our method consistently outperforms or matches online fine-tuning baselines and achieves significant improvements over the previous action preference query approach.

\begin{ack}
This work was supported by the NSFC(U2441285, 62222605).
\end{ack}
\bibliography{mybibfile}

\end{document}